\documentclass[10pt,twocolumn,letterpaper]{article}

\usepackage{cvpr}              
\usepackage{xcolor}
\usepackage{listings}
\usepackage{comment}
\usepackage[T1]{fontenc}
\usepackage{textcomp}
\definecolor{jsonkey}{RGB}{150,40,120}
\definecolor{jsonstring}{RGB}{20,110,60}
\definecolor{jsonpunct}{RGB}{50,80,160}

\lstdefinestyle{coloredjson}{
    basicstyle=\ttfamily\tiny\color{jsonpunct},
    breaklines=true,
    numbers=left,
    numberstyle=\tiny\color{gray},
    numbersep=6pt,
    frame=single,
    rulecolor=\color{gray!50},
    backgroundcolor=\color{gray!5},
    keepspaces=true,
    columns=fullflexible,
    showstringspaces=false,
    string=[s]{"}{"},
    stringstyle=\color{jsonstring},
    moredelim=**[s][\color{jsonkey}]{"type"}{:},
    moredelim=**[s][\color{jsonkey}]{"properties"}{:},
    moredelim=**[s][\color{jsonkey}]{"enum"}{:},
    moredelim=**[s][\color{jsonkey}]{"description"}{:},
    moredelim=**[s][\color{jsonkey}]{"format"}{:},
    moredelim=**[s][\color{jsonkey}]{"required"}{:},
    moredelim=**[s][\color{jsonkey}]{"sky_condition"}{:},
    moredelim=**[s][\color{jsonkey}]{"precipitation"}{:},
    moredelim=**[s][\color{jsonkey}]{"lighting_condition"}{:},
    moredelim=**[s][\color{jsonkey}]{"visibility"}{:},
    moredelim=**[s][\color{jsonkey}]{"surface_condition"}{:},
    moredelim=**[s][\color{jsonkey}]{"wind_effects"}{:},
    moredelim=**[s][\color{jsonkey}]{"dominant_weather_effect"}{:},
    moredelim=**[s][\color{jsonkey}]{"confidence_score"}{:},
    moredelim=**[s][\color{jsonkey}]{"reasoning"}{:}
}
\usepackage{tabularx}

\usepackage{url}

\usepackage{minted} 

\usepackage{multirow}
\usepackage{makecell}
\usepackage{dsfont}
\usepackage{mathrsfs}
\usepackage[accsupp]{axessibility}
\usepackage{subcaption}
\usepackage{float} 

\definecolor{cvprblue}{rgb}{0.21,0.49,0.74}
\usepackage[pagebackref,breaklinks,colorlinks,allcolors=cvprblue]{hyperref}

\def\paperID{19} 
\def\confName{CVPR}
\def\confYear{2026}

\title{Structure-retained low-rank adapters for weather synthesis }

\author{Shunxin Wang\footnotemark \and Alexandros Stergiou \and Luuk Spreeuwers \and Estefania Talavera \and  Nicola Strisciuglio \\
{\tt\small shunxinwang1030@gmail.com}
}

\begin{document}
\maketitle
\begin{abstract}
Real-world deployment of computer vision models, particularly in outdoor scenarios, is often limited by environmental interferences, such as adverse weather. Improving model robustness across weather conditions is thus critical for reliability. However, sourcing high-quality data relevant to isolated weather effects is a significant challenge. Most datasets include changes in dynamic elements across observations of the same scene: subjects such as pedestrians and vehicles change between shots, while camera viewpoints vary. These aspects make it difficult to decouple solely weather-induced distribution shifts from the scene. In this work, we address this with a novel weather synthesis framework that utilizes structure-retaining Low-Rank Adapters (LoRA). We analyze the attention blocks in Stable Diffusion XL (SDXL) and identify where relevant structural information is mostly processed during image-to-image translation. Through an architectural analysis of SDXL, we learn structural LoRA matrices for the corresponding blocks and then leverage them for weather synthesis, eliminating the need for external conditioning such as ControlNet. Experimental results demonstrate that our method improves upon existing synthesis approaches, achieving greater realism while preserving structure. This provides a robust pipeline for generating high-fidelity training and evaluation data for outdoor computer vision applications.
\end{abstract}  

\let\thefootnote\relax\footnotetext{$^*$work done while at the University of Twente}
\section{Introduction}
\label{sec:intro}
Computer vision models deployed in real-world scenarios often encounter robustness problems. In outdoor settings, interference comes not only from hardware, such as photon and thermal noise (especially in low-light settings), but also from unpredictable environmental changes, such as weather conditions. Improving model performance is important for ensuring the reliability of deployed models, especially in safety-critical applications such as autonomous driving. However, collecting high-quality, sufficiently varied image data that contains various weather changes remains a challenge. Sourcing real-world image data from the same scene in different weather conditions requires comprehensive control over dynamic objects. Existing datasets mostly contain image pairs under different conditions, e.g., sunny-rainy and sunny-foggy, without ensuring that the original objects present in the images (e.g., cars and passengers) or camera views are preserved~\cite{9318521,ijcai2022p243}. This induces not only weather-related distribution shifts, but also temporal and spatial variations.

Generative models can produce high-quality images while offering granular control over details like style (e.g., weather) and specific scene content~\cite{poole2023dreamfusion,ruiz2023dreamboothfinetuningtexttoimage}.
For style manipulation, previous works extracted style information from one reference image~\cite{jeong2024visualstylepromptingswapping}, or from a pair of images with and without the target style~\cite{jones2024customizingtexttoimagemodelssingle}. For text-to-image generation tasks, image style can be manipulated by changing style-related text descriptions. However, for image-to-image translation, manipulating the input image style using only text prompts is insufficient, either failing to produce strong changes or disrupting the image content~\cite{WU2024103887,10100797830319780_41}.

\begin{figure}[!t]
    \centering
    \includegraphics[width=\linewidth]{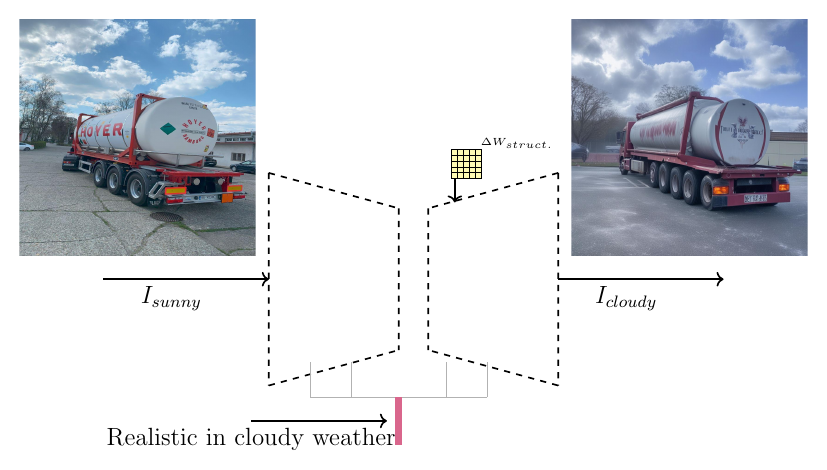}
    \caption{Text-guided weather manipulation with structure-LoRA for preserving the structural layout of input images.  }
    \label{fig:teaser}
\end{figure}

Synthesizing adverse weather conditions is a critical frontier for improving computer vision models deployed in outdoor scenarios, e.g., autonomous driving. 
Previous work relied on Generative Adversarial Networks (GANs), such as CycleGAN~\cite{zhu2020unpairedimagetoimagetranslationusing} that simulated diverse weather types. These methods often require high-quality paired training data, which is expensive to obtain. To circumvent this, ~\cite{lin2025controllableweathersynthesisremoval} turned to graphics engines and large-scale generative models. For instance, Stable Diffusion XL (SDXL)~\cite{podell2023sdxlimprovinglatentdiffusion} is used to curate precisely controlled weather attributes within synthetic scenes. 
Despite the increased variety these models offer, the original scene structure is often not preserved during image-to-image translation.
ControlNet~\cite{zhang2023addingconditionalcontroltexttoimage} was proposed to preserve structure information of an input image, by injecting structured spatial conditions, e.g., edges and depth, into intermediate feature maps. 
But this often leads to a loss of fine-grained details, such as original textures and object identities~\cite{huang2024controllable}. Other approaches that adopt instruction-based editing, such as InstructPix2Pix~\cite{brooks2023instructpix2pixlearningfollowimage}, frequently struggle with spatial relational reasoning, i.e., the models either focus solely on local effects (e.g., rain in the sky) while neglecting global consequences (e.g., reflections on the road), or alter global image color rather than specific weather elements.

In this work, we perform an architectural analysis of SDXL on the sensitivity of attention blocks to structural information, and subsequently devise a structure-retaining Low-Rank Adaptation (LoRA)~\cite{hu2021loralowrankadaptationlarge} strategy to synthesize weather while preserving structure details. 
Compared to other weather synthesis methods, our method improves the variety of weather effects while requiring light computations. Compared with other style-manipulation approaches, our method preserves more structural and content information. Effectively, we reduce the need to curate high-quality paired image data through manual collection or graphic engines. 

Our contributions are as follows:
\begin{enumerate}
    \item An architectural analysis and identification of SDXL attention blocks that capture more structural and dynamic information within scenes;
    \item A weather synthesis framework using structural priors in the form of low-rank adapters (structure-LoRA), to retain the layout and dynamic elements of input images. Our method improves realism and weather-synthesis effects while better preserving scene dynamic objects and structural layout compared to existing approaches. 
\end{enumerate}

\section{Related works}
\paragraph{Style transfer}
Image style transfer aims to change the style of a source image to match that of a reference image.
With the development of vision-language models (VLMs), works have leveraged encoded semantic knowledge. In~\cite{hertz2022prompttopromptimageeditingcross}, it was shown that cross-attention layers in text-conditioned diffusion models extract spatial relationships between image pixels and prompt tokens. Manipulating cross-attention maps by changing only the text prompts (e.g., replacing a word or adding a specification) allows editing an image while preserving most of the original structural information. Similarly, the utility of self-attention for capturing spatial relationships (e.g., color layouts) and local visual features (e.g., textures)was investigated in~\cite{jeong2024visualstylepromptingswapping}. Replacing attention matrices of style reference images with those of a content image can preserve the original content while manipulating the style of the original image. However, some content of the style reference image might leak to the generated images, since content and style information extracted in the attention layers are entangled. By enabling `attention sharing’ during the diffusion process, style-consistent images with the first (i.e., style reference) image in a batch were generated, without optimization~\cite{hertz2024stylealignedimagegeneration} . Explicitly learning style information, InST~\cite{zhang2023inversionbasedstyletransferdiffusion} maps the style representation of a reference image to text tokens that do not exist in the normal language, e.g., \texttt{[C]}, and then provides such a text prompt to manipulate the style of the input images. 
Instead of text prompts as conditions, IP-Adapter~\cite{ye2023ipadaptertextcompatibleimage} additionally injected image information extracted from a style/structure reference image into the attention layers of diffusion models. However, this approach is limited to either style manipulation or structure generation. For text-to-image (T2I) style transfer, the authors of~\cite{chen2024artadaptertexttoimagestyletransfer} used a style encoder to extract multi-level style embeddings aligned with textual descriptions alongside an auxiliary content adapter to separate content and style.

To manipulate style information more precisely, several works have focused on decomposing style and content. This is done by disentangling and then replacing the style or content components of the images: the generated images may change in subject or style~\cite{xu2024breakformakemodularlowrankadaptations,wang2024instantstyle}. The authors of~\cite{frenkel2024implicitstylecontentseparationusing} explored the sensitivity of each block of StableDiffusion XL to  style or content information. This analysis guides the implicit separation of a single image style and content components by jointly learning the LoRA weights of the content-sensitive and style-sensitive blocks. UnZipLoRA~\cite{liu2025unziploraseparatingcontentstyle} disentangles a single stylized image into content and style information as two independent LoRAs by encouraging orthogonality through regularization of LoRA weight matrices. PairCustomization~\cite{jones2024customizingtexttoimagemodelssingle} learns the stylistic differences from an image pair (a content image and a style reference image with similar structure) by jointly optimizing content-LoRA to reconstruct the content image and the style-LoRA to edit the style of the content image. However, these methods require per-pair or per-image optimization, which can be computationally expensive and impractical for real-time inference. The work of~\cite{shenaj2025lorararlearningmergeloras} introduces a hypernetwork that predicts the merging coefficients of content and style LoRAs, which generalizes well to unseen content and style LoRA pairs, thus eliminating the need for costly optimization. CSGO~\cite{xing2024csgocontentstylecompositiontexttoimage} trains a style-transfer model that takes style and content images as inputs, and learns to fuse content and style features by a feature injection block. This allows for style transfer during inference without fine-tuning or optimization.


Rather than learning LoRA weights for style and content information, the work of~\cite{choi2025stylefriendlysnrsamplerstyledriven} found that using the same noise sampling distributions and loss weighting schedules as those in the original model's training might not address style features during diffusion. Namely, stylistic features are more prominent in the early denoising steps, which have a low signal-to-noise ratio, than in later steps. Based on this, ~\cite{choi2025stylefriendlysnrsamplerstyledriven} shifted the mean of the noise sampling distribution to induce more noise during fine-tuning, allowing for better style alignment. 

Our structure-LoRA eliminates the need for a style reference image and modulates the weather of original images using only text prompts, thus avoiding content leakage. Importantly, structure-LoRA preserves the structure and content of input images, whereas existing approaches to style manipulation either require additional structural conditions or do not address structure preservation altogether.

\paragraph{Weather synthesis}
Adverse weather synthesis is important for computer vision applications, such as autonomous driving~\cite{Musat_2021_ICCV}. It facilitates evaluating model robustness under varying weather conditions and improving vision models in adverse weather. In~\cite{Musat_2021_ICCV} and~\cite{zhu2020unpairedimagetoimagetranslationusing}, GANs and CycleGANs were applied respectively, to generate images with 14 adverse weather conditions. The work of~\cite{tremblay2021rain} uses both physics-based and GAN-based methods to augment data with rain effects. However, these methods need high-quality paired image data for model training. The work of~\cite{lin2025controllableweathersynthesisremoval} curated data using graphics engines that precisely control weather attributes of the same scene, as well as a large image generative model such as StableDiffusion XL~\cite{podell2023sdxlimprovinglatentdiffusion}. 
With the development of generative vision-language models, the authors of~\cite{Marathe_2023_CVPR} leveraged DALL-E~\cite{ramesh2021zeroshottexttoimagegeneration} to synthesize weather effects given pre-designed text prompts. 
To preserve the structural layout of images, the work of~\cite{10959647,loiseau2024reliabilitysemanticsegmentationuse} used ControlNet~\cite{zhang2023addingconditionalcontroltexttoimage}, which takes structure-condition images, e.g., Canny edge images. However, many precise details may be lost during the generative process, such as textures, color, and identity. 
Rather than using ControlNet, \cite{10927434,10870179} generated weather-modulated images with instruction editing models, such as  InstructPix2Pix\cite{brooks2023instructpix2pixlearningfollowimage}. These models are trained on well-crafted synthetic image pairs with diverse and extreme weather conditions generated by advanced conditional diffusion models. However, local details of input images in inference time might not be preserved. 


\section{Preliminary}

We investigate how changes to the condition embeddings affect the structural layout of the generated images. 

\noindent
\textbf{SDXL}~\cite{podell2023sdxlimprovinglatentdiffusion} consists of a 70-attention block UNet, each with cross- and self-attention blocks. SDXL is trained on a latent diffusion objective ${\mathcal{L}_{LDM} = \mathbb{E}_{z \sim \mathcal{E}(x), \epsilon \sim \mathcal{N}(0,1), t} [ \| \epsilon - \epsilon_\theta(z_t, t, c) \|_2^2 ]}$, where $\epsilon_\theta$ predicts the noise $\epsilon\sim \mathcal{N}(0,1)$ added at timestep $t$ to latents $\mathbf{z} \in \mathbb{R}^{H/8 \times W/8 \times 4}$ corresponding to image $\mathbf{x} \in \mathbb{R}^{H \times W \times 3}$ encoded with $\mathcal{E}(\cdot)$.

\begin{figure*}
    \centering
    \subfloat[]{\includegraphics[width=0.33\linewidth]{ 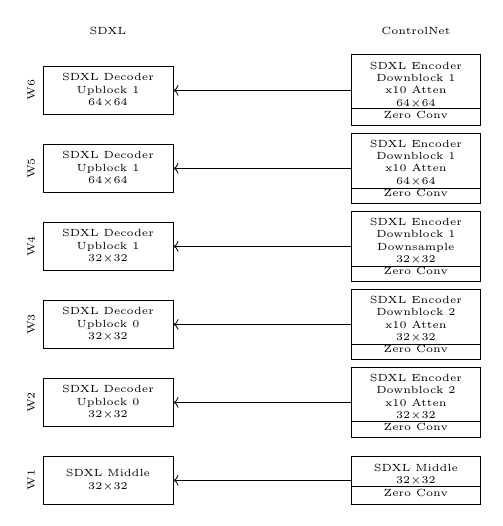}\label{architecture}} 
    \subfloat[]{\includegraphics[width=0.33\linewidth]{ 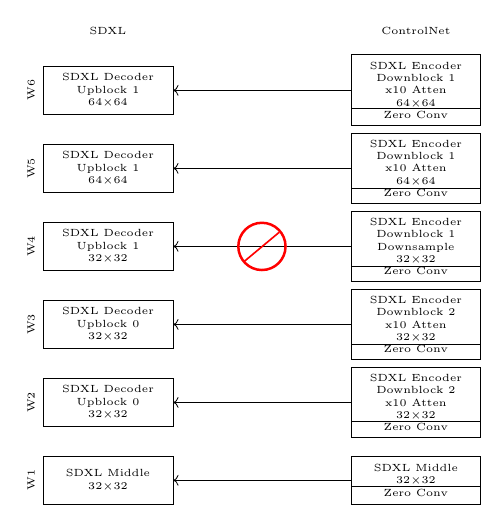}}
    \subfloat[]{\includegraphics[width=0.33\linewidth]{ 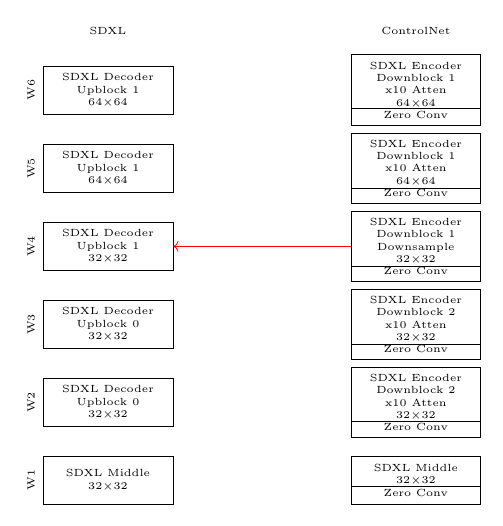}}
    \caption{ (a) Partial SDXL-ControlNet architecture. We use two strategies: (b) removing one of the embeddings and (c) retaining only one embedding to examine the influence of the condition embeddings fed to different blocks of SDXL on the structural information of the generated images. The weights of each block are denoted as $W{i}$. }
    \label{fig:maskout_strg}
    \vspace{-1em}
\end{figure*}

\noindent
\textbf{ControlNet}~\cite{zhang2023addingconditionalcontroltexttoimage} creates a trainable copy $\theta'$ of the UNet encoder part of SDXL, keeping parameters frozen. Zero convolutions, $1\times1$ convolutions with zero-init weights and biases are added between the frozen and trainable copy. For a given UNet block $\mathcal{F_\theta}$, the output $y_c$ is formulated as $y_c = \mathcal{F_\theta}(\mathbf{z}_F) + \mathcal{Z}(\mathcal{F_{\theta'}}(\mathbf{z}_F + \mathbf{c}))$,
where $\mathbf{z}_F \in \mathbb{R}^{h \times w \times d}$ is the input feature map and  $\mathbf{c} \in \mathbb{R}^{h \times w \times d}$ is the added condition after preprocessing. 
We show the partial view of SDXL with ControlNet in~\cref{architecture}.


\noindent
\textbf{LoRA.} Low-rank adapters are efficient modules that enable fine-tuning for large pre-trained models by optimizing rank decomposition matrices~\cite{hu2021loralowrankadaptationlarge}. Given the weights of a pre-trained text-to-image diffusion model ($W_0 \in \mathbb{R}^{d\times k}$), LoRA learns an update matrix $\Delta W$ that can be decomposed into two low intrinsic rank matrices $A\in \mathbb{R}^{r\times k}$ and $B\in \mathbb{R}^{d\times r}$, with the rank $r \ll min(d,k)$ (i.e., $\Delta W= BA)$. During training, $W_0$ is frozen, while the two low intrinsic rank matrices $A$ and $B$ are updated. During inference, the model weights are fine-tuned by $W = W_0 + \Delta W$. This approach promotes efficiency by restricting the number of tunable parameters. 

\section{Method}

\subsection{SDXL block structure priors injection}
Prior research examined how manipulating style or content keywords in text prompts, injected into various UNet blocks, impacts the resulting visual style and content~\cite{frenkel2024implicitstylecontentseparationusing}.
However, sole text prompt manipulation through injection and replacement does not expose effects on structural image information. 

In this work, we use ControlNet on SDXL to inject structural priors to each SDXL block. The combined model then generates structure-preserved images guided by text. 
Following an approach similar to \cite{frenkel2024implicitstylecontentseparationusing}, we investigate how manipulating ControlNet-encoded condition embeddings that are injected into SDXL affects the generated structure information. Specifically, we: (1) mask one of the condition paths while keeping the others unchanged; (2) maintain only one condition path, removing all others. 
We replace the masked path(s) with zero values to serve as \emph{null conditions} (see~\cref{fig:maskout_strg}).
During image generation, we employ a prompt template \texttt{A photo of a [category name]} alongside ControlNet that uses the Canny edges of the source image as the additional condition. 

\noindent
\textbf{Dataset.} To visualize and quantify how structural information is affected, we selected images from the Caltech 256 dataset~\cite{griffin_holub_perona_2022}, with simple backgrounds and clear object contours\footnote{The classes are: \texttt{human skeleton}, \texttt{helicopter}, \texttt{bathtub}, \texttt{tuning fork}, \texttt{T-shirt}, \texttt{ladder}, \texttt{fighter jet}, \texttt{revolver}, \texttt{bowling pin}, \texttt{cormorant}, \texttt{spoon}, \texttt{tricycle}, \texttt{knift}, \texttt{frisbee}, \texttt{sword}, \texttt{eyeglasses}, \texttt{diamond ring}, \texttt{tripod ipod}, \texttt{teadpot}, \texttt{iguana}, \texttt{yoyo}, \texttt{soccer ball}, \texttt{ak47}, and \texttt{penguin}.}. 



\begin{figure*}[!t]
    \centering
    \subfloat[]{ \includegraphics[width=.48\linewidth]{ 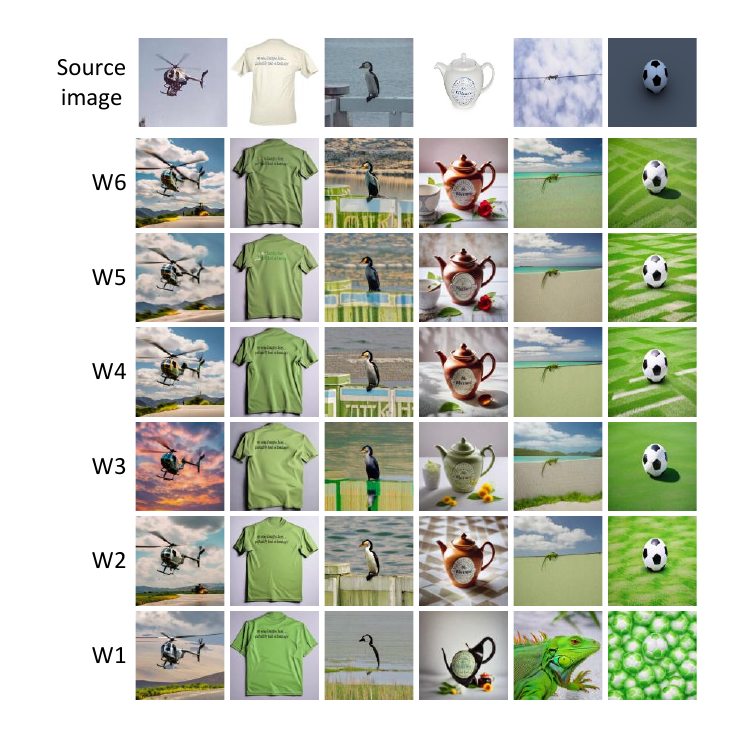}\vspace{-1mm}\label{strg1}}
    \subfloat[]{\includegraphics[width =.48\linewidth]{ 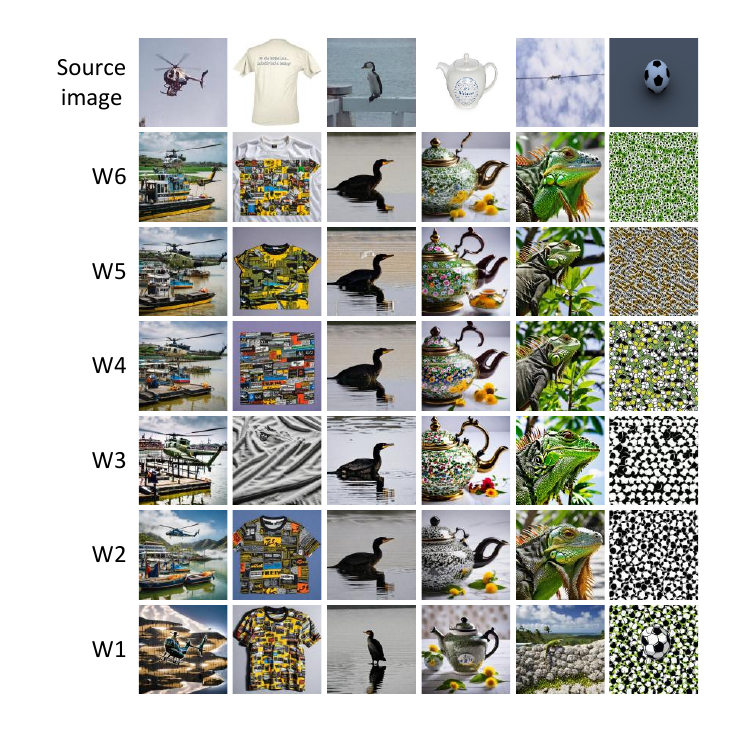}\vspace{-1mm}\label{strg2}}
    \caption{Qualitative results analyzing SDXL architecture using conditional embeddings masking out (a) strategy 1  and (b) strategy 2. }
    \label{fig:structAnalysis}
\end{figure*}

\noindent
\textbf{Visualization results.} Masking out the condition embedding fed to the middle block (noted as $W_1$) of SDXL results in images that do not preserve the original structure. As shown in~\cref{strg1}, main objects either disappear, such as the cormorant, or become heavily distorted, such as the unrecognizable teapot or the zoomed-in iguana. Similarly, we observe from~\cref{strg2} that retaining only the condition from the middle block enables SDXL to recover most structural details. Thus, for structure-preserving weather synthesis, we target optimizing the LoRA weights of the middle block of SDXL.

\subsection{Structure-LoRA for weather synthesis}
\begin{figure}[!t]
    \centering
    \includegraphics[width=\linewidth]{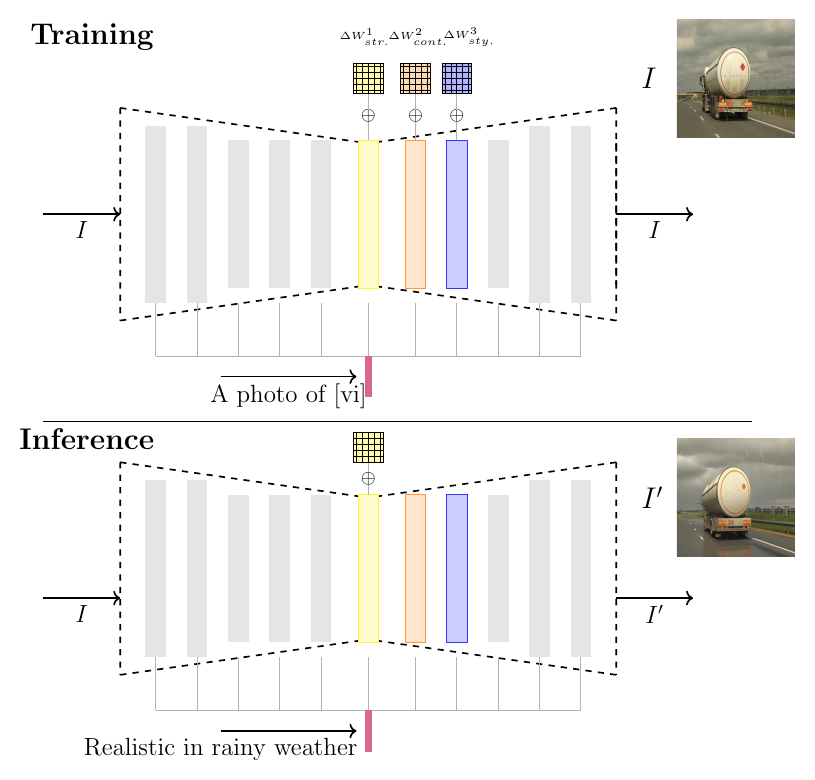}
    \caption{Learning LoRA matrices for structure, style and content information separately. During inference, we apply the structure LoRA matrix for preserving the structural layout of images.}
    \vspace{-2mm}
    \label{fig:schema}
\end{figure}

Guided by these insights, we design a structure-enhanced weather-synthesis adapter, offering a more lightweight alternative to traditional ControlNet-based conditioning. We follow DreamBooth LoRA~\cite{ruiz2023dreamboothfinetuningtexttoimage} and finetune SDXL on a standard reconstruction objective using a general text prompt \texttt{A photo of [$v_i$]}. We avoid using \texttt{A photo of a [category name]}, since prompts with category names strongly bias the image generation towards the specific category, reducing the utility of the learned LoRAs.
We follow the optimization setup of~\cite{frenkel2024implicitstylecontentseparationusing} to learn different LoRAs to represent varying styles, content, and structural information. At inference, a prompt template \texttt{realistic in [rainy/cloudy/snowy/foggy] weather} is used alongside the source image to guide the generation towards target weather effects, using a specific LoRA weight; e.g., structure (struct.) LoRA, content (cont.) LoRA, style (sty.) LoRA, or a combination of them. We overview the pipeline in~\cref{fig:schema}.

\paragraph{Metrics.} We use the Structure Similarity Index Measure (SSIM)~\cite{1284395} and the Learned Perceptual Image Patch Similarity (LPIPS)~\cite{zhang2018perceptual} to evaluate the preservation and realism of the structure of the generated images. SSIM quantifies the structural fidelity between source and generated images, ranging from 0 to 1, with higher values indicating better structure preservation. LPIPS measures perceptual differences aligned with human visual judgment using deep network feature embeddings. 
Lower LPIPS scores indicate greater perceptual similarity to human observation and thus better image quality and realism. 

To evaluate the success rate of weather effect synthesis, we deploy an \textit{LLM-as-judge} approach, using the prediction output of Gemini 2.5 flash. Giving a text prompt \texttt{You are an image analysis model specializing in meteorological observations. Your task is to detect and describe visible weather effects in the provided image. Respond STRICTLY according to the JSON schema provided in Listing \ref{lst:weather-json}}. The detected dominant weather effect is taken as the prediction of the model.

\section{Experiments}

\noindent
\textbf{Baselines and training settings.} We select B-LoRA~\cite{frenkel2024implicitstylecontentseparationusing} and PairCustomization~\cite{jones2024customizingtexttoimagemodelssingle} as baselines. These two parameter-efficient fine-tuning methods are designed for style manipulation using style and content disentangled LoRA weights, offering a straightforward comparison to our structure-LoRA regarding structural preservation.   
B-LoRA and PairCustomization can generate images in two setups, which are (1) text-to-image generation (T2I) and (2) image-to-image generation (I2I). Our method uses I2I generation to preserve structural information as much as possible, while manipulating the weather style with fewer constraints from the original image input. For a comprehensive analysis of the performance, we compare our method to the baseline methods using both settings. During I2I, the hyperparameter `strength' controls the guidance scale of text prompts. A lower strength indicates details of an input image are largely preserved, while a higher strength indicates significant changes applied to the input image guided by the text prompt. 

\begin{figure}[!t]
    \centering
    \subfloat[]{%
        \includegraphics[height=0.25\textwidth]{ 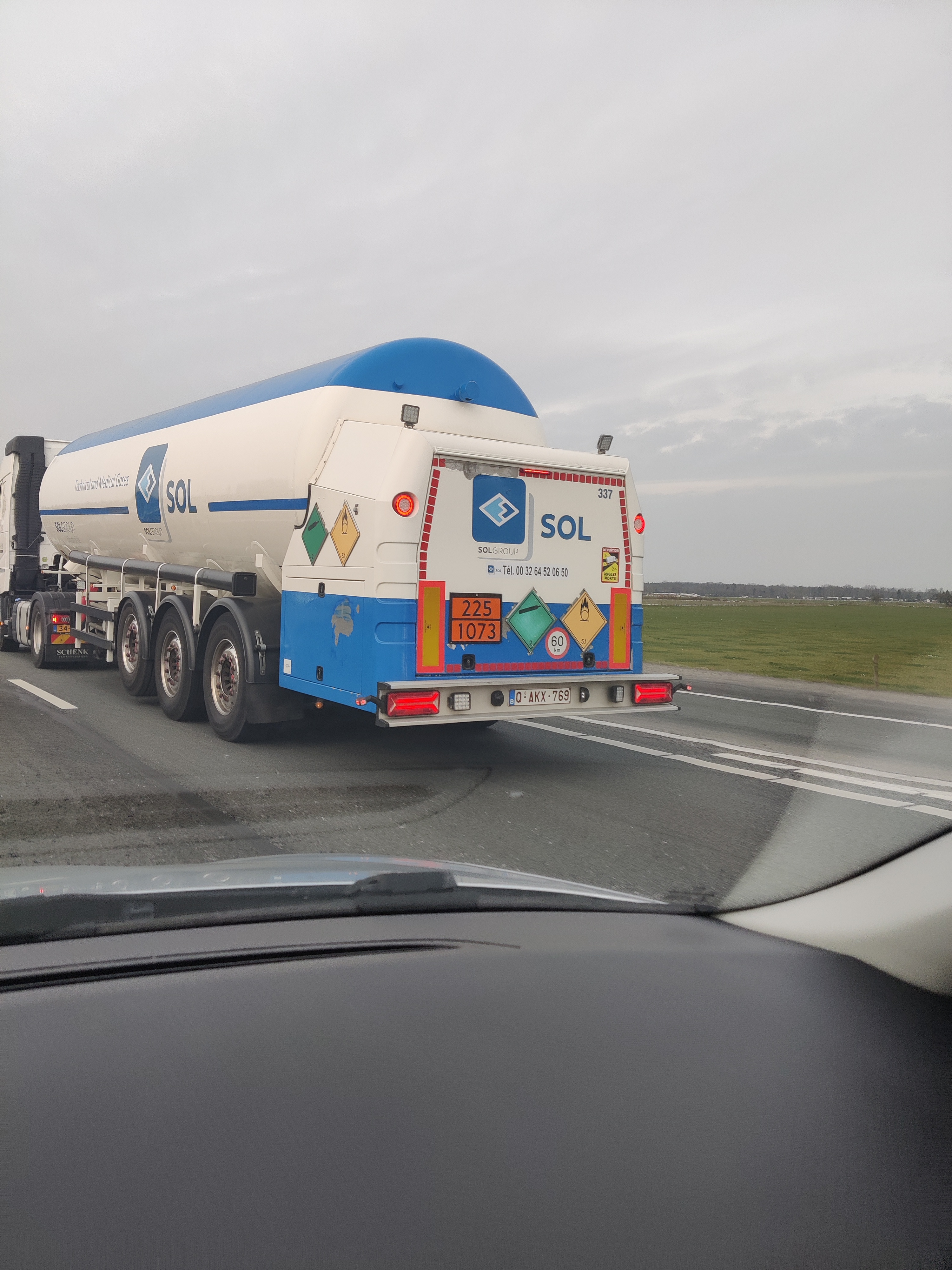}%
        \label{fig:image1}
    }    
    \subfloat[]{%
        \includegraphics[height=0.25\textwidth]{ 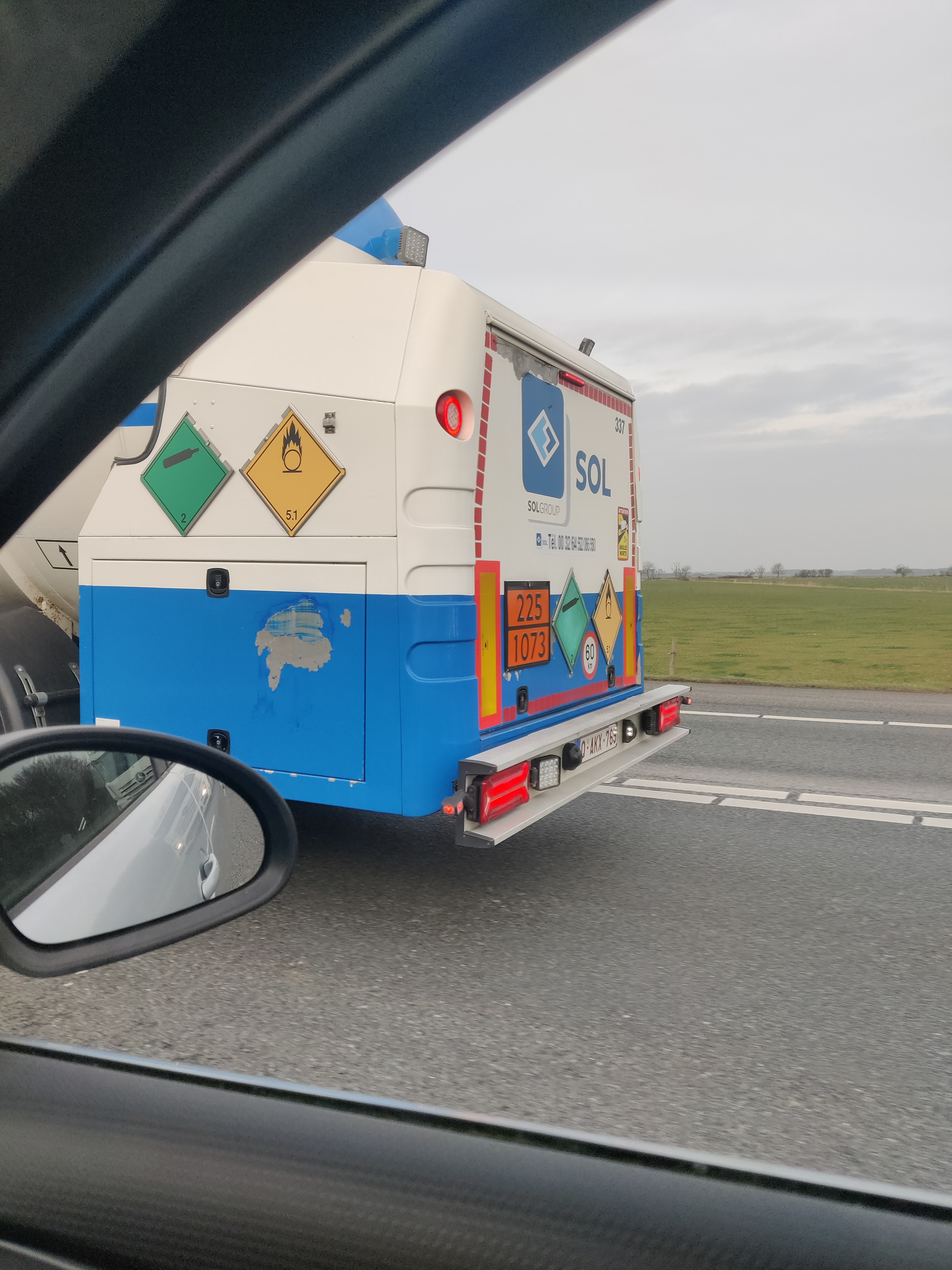}%
        \label{fig:image2}
    }
    \caption{Example images from Haztruck dataset.}
    \label{fig:Haztrucksamples}
\end{figure}

\begin{lstlisting}[style=coloredjson,
    float=tbp,
    caption={JSON schema for weather analysis.},
    label={lst:weather-json},
    captionpos=b]
{
    "type": "OBJECT",
    "properties": {
        "sky_condition": {
            "type": "STRING",
            "enum": ["clear", "partly_cloudy", "overcast", "stormy", "foggy", "hazy"],
            "description": "The dominant sky condition."
        },
        "precipitation": {
            "type": "STRING",
            "enum": ["none", "rain", "snow", "hail", "sleet", "drizzle"],
            "description": "The type of precipitation observed."
        },
        "lighting_condition": {
            "type": "STRING",
            "enum": ["bright", "dim", "sunset", "night", "artificial"],
            "description": "The observed lighting condition."
        },
        "visibility": {
            "type": "STRING",
            "enum": ["good", "moderate", "poor"],
            "description": "The general visibility level."
        },
        "surface_condition": {
            "type": "STRING",
            "enum": ["dry", "wet", "snowy", "icy", "muddy"],
            "description": "The condition of the ground surface."
        },
        "wind_effects": {
            "type": "STRING",
            "enum": ["none", "visible motion blur", "bent trees", "blowing debris"],
            "description": "Observable signs of wind."
        },
        "dominant_weather_effect": {
            "type": "STRING",
            "enum": ["sunny", "cloudy", "rainy", "snowy", "foggy", "stormy", "other"],
            "description": "The single most dominant weather phenomenon."
        },
        "confidence_score": {
            "type": "NUMBER",
            "format": "float",
            "description": "Confidence of the analysis, between 0.0 and 1.0."
        },
        "reasoning": {
            "type": "STRING",
            "description": "A brief explanation of the visual cues used for the analysis."
        }
    },
    "required": [
        "sky_condition", "precipitation", "lighting_condition", "visibility",
        "surface_condition", "wind_effects", "dominant_weather_effect",
        "confidence_score", "reasoning"
    ]
}
\end{lstlisting}

In our baseline experiments, we learn the weather style LoRA from a reference image depicting rain, snow, fog, or clouds.  
B-LoRA extracts style LoRA weights from a reference image or a set of similarly styled reference images. PairCustomization learns style-LoRA and content-LoRA from the pairwise differences between images. For weather conditions, this can include \texttt{sunny} - [\texttt{rainy}/\texttt{snowy}/\texttt{foggy}/\texttt{cloudy}] image pairs in exactly the same scene. Natural image datasets containing such image pairs are hard to obtain. We thus use Gemini 2.5 Flash Image (Nano Banana)~\cite{fortin2025introducing} to generate these image pairs with text prompts modifying only the weather description.

\noindent
\textbf{Dataset.}
We select haztruck dataset, which contains  truck images with hazmat plates, sourced from the internet~\cite{1010079783032049681_22}. The dataset is curated for hazmat plate detection and optical character recognition. After filtering low-quality images, we use 174 images out of 238 images. Example images are shown in~\cref{fig:Haztrucksamples}.

\subsection{Quantitative results}
\begin{figure}[h]
    \centering
    \hspace{-2.5mm}
    \subfloat[]{ \includegraphics[width=0.497\linewidth]{ 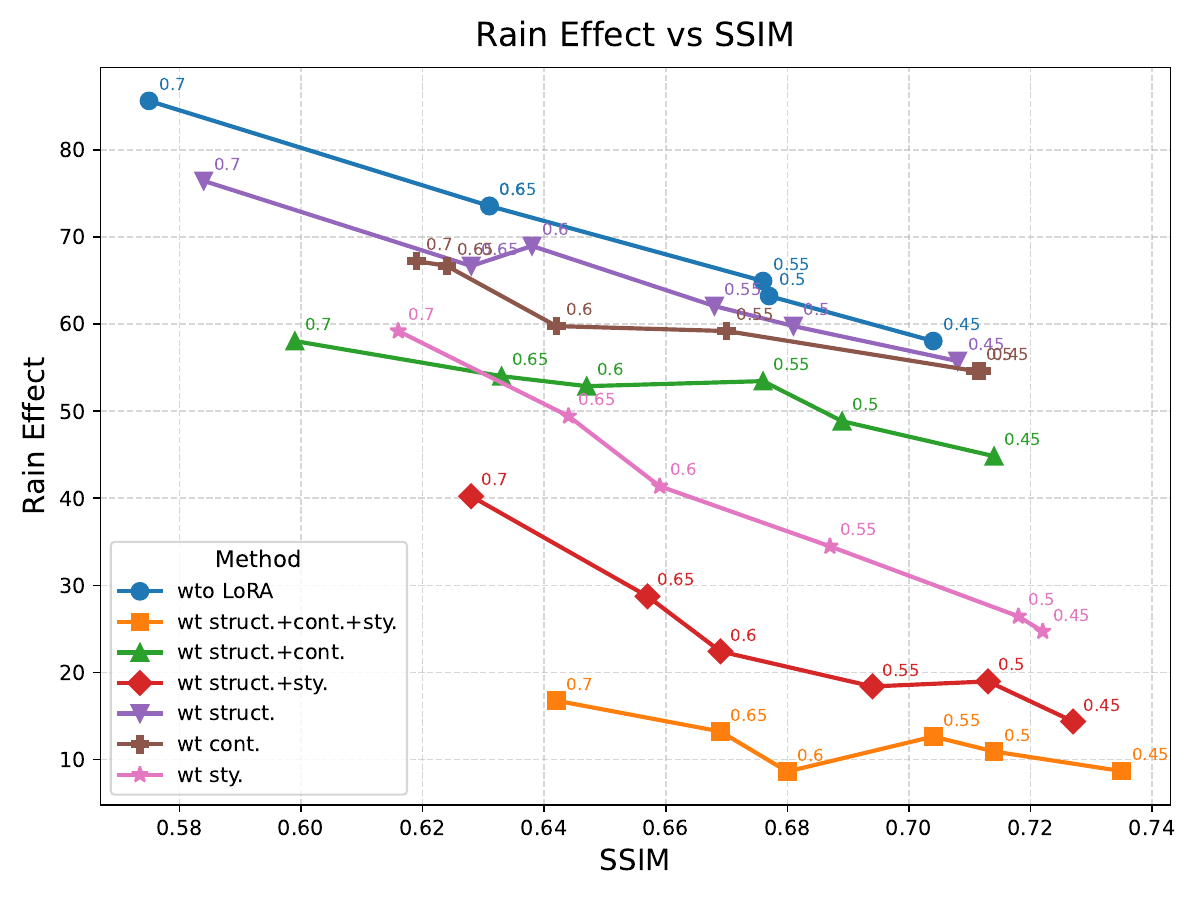}
    }
    \subfloat[]{\includegraphics[width=0.497\linewidth]{ 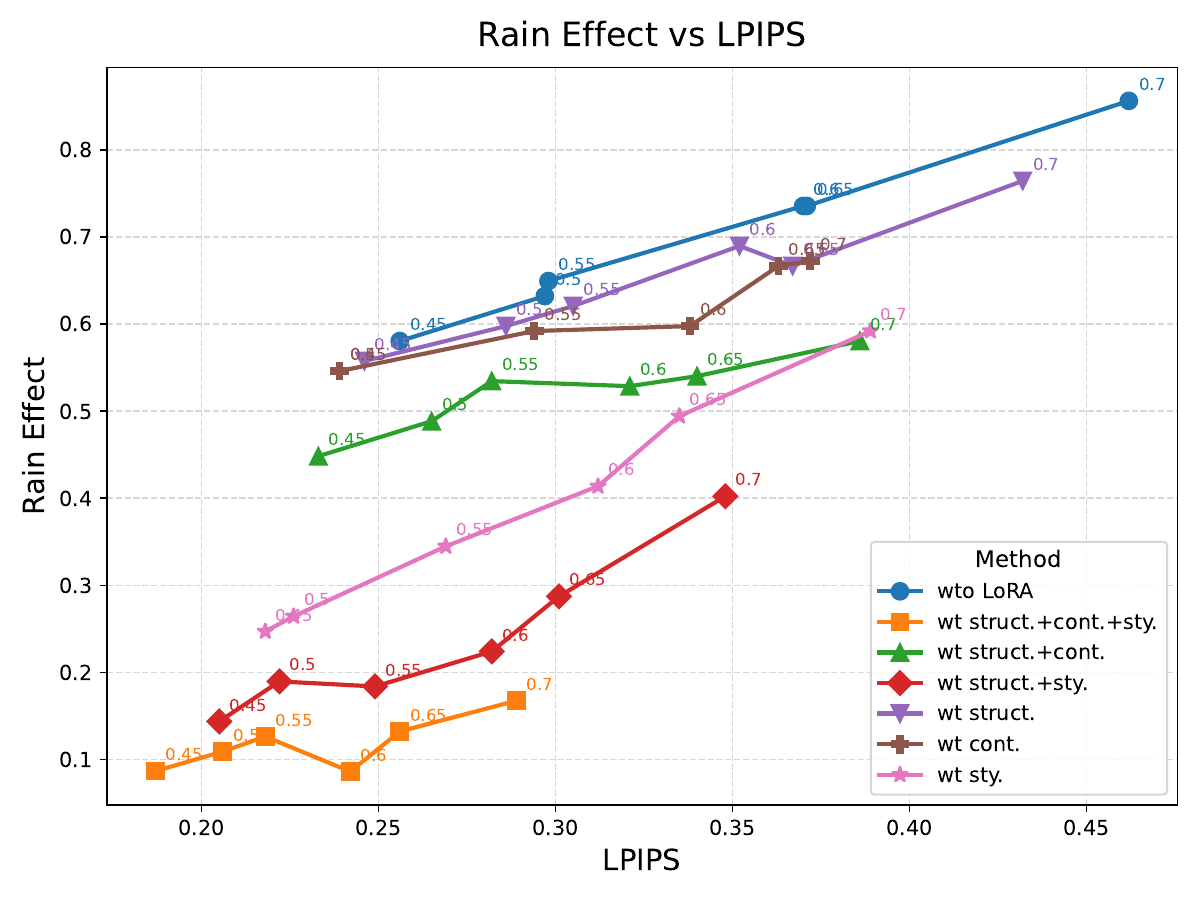}}

    \subfloat[]{\includegraphics[width=0.5\linewidth]{ 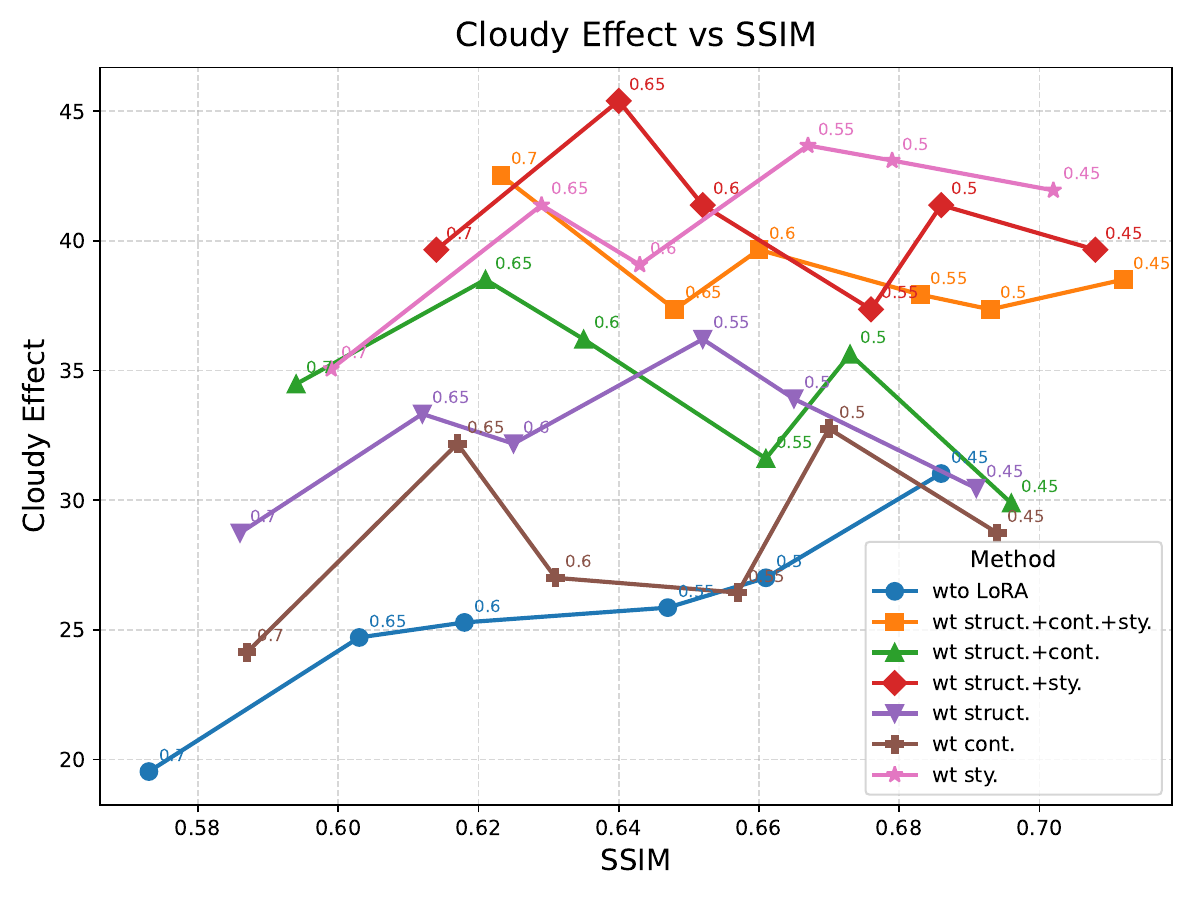}}
    \subfloat[]{\includegraphics[width=0.5\linewidth]{ 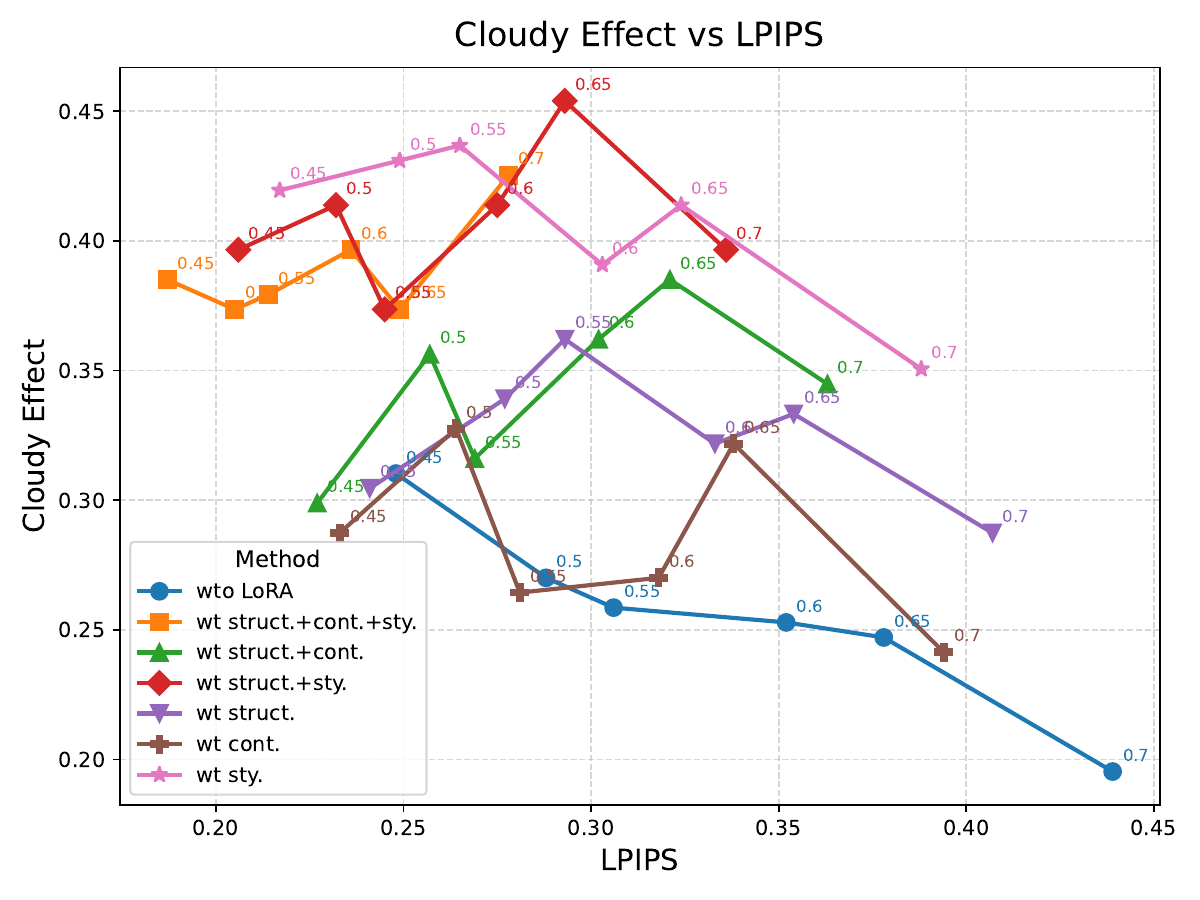}}   
    
    \subfloat[]{\includegraphics[width=0.5\linewidth]{ 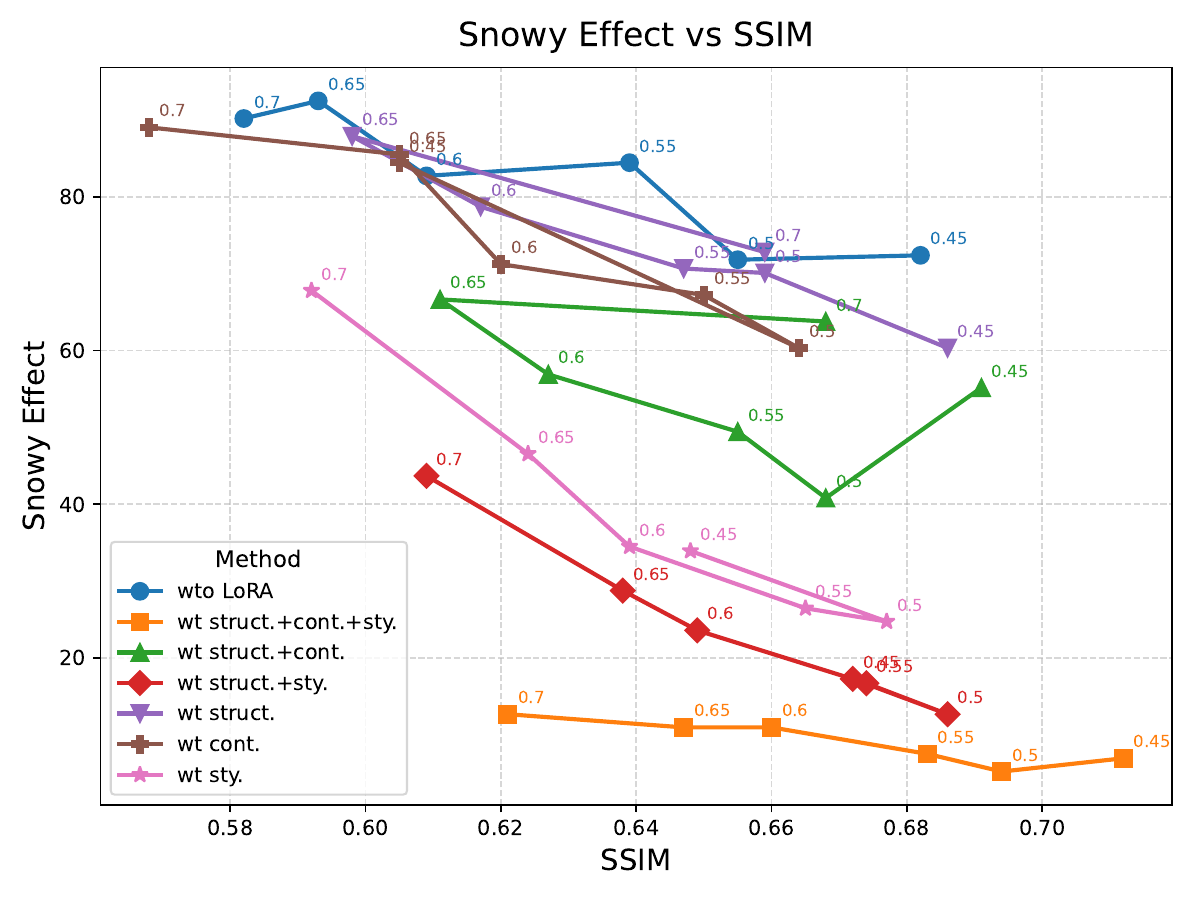}}
    \subfloat[]{\includegraphics[width=0.5\linewidth]{ 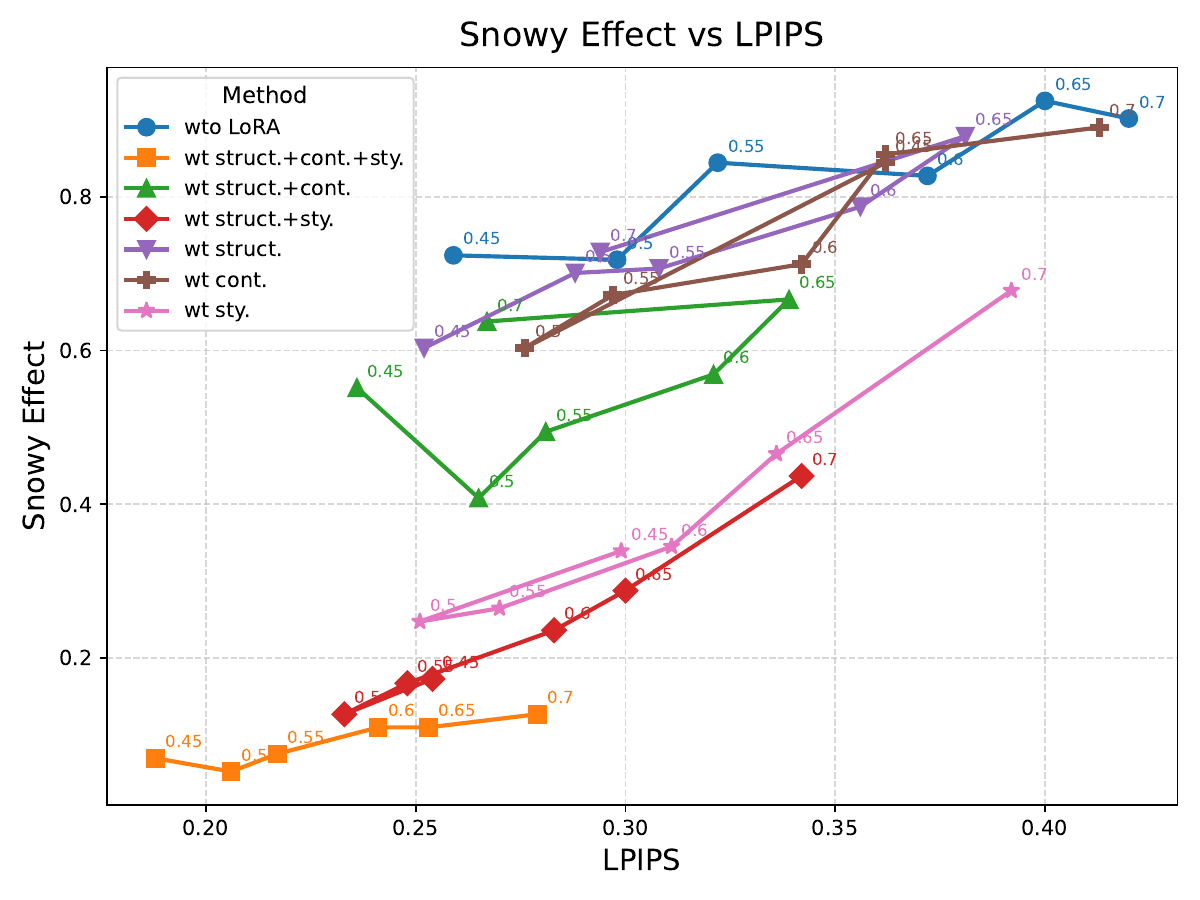}}
    
    \subfloat[]{\includegraphics[width=0.5\linewidth]{ 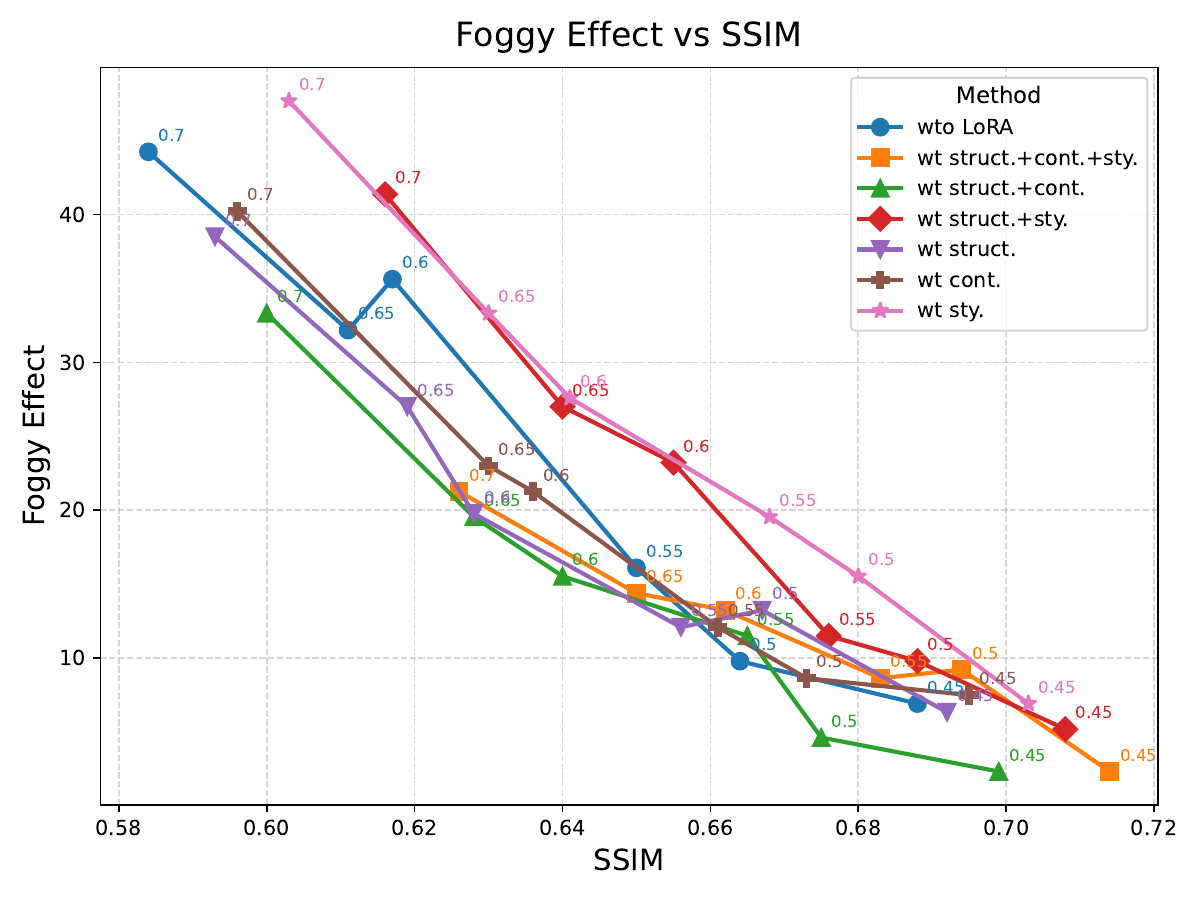}}
    \subfloat[]{\includegraphics[width=0.5\linewidth]{ 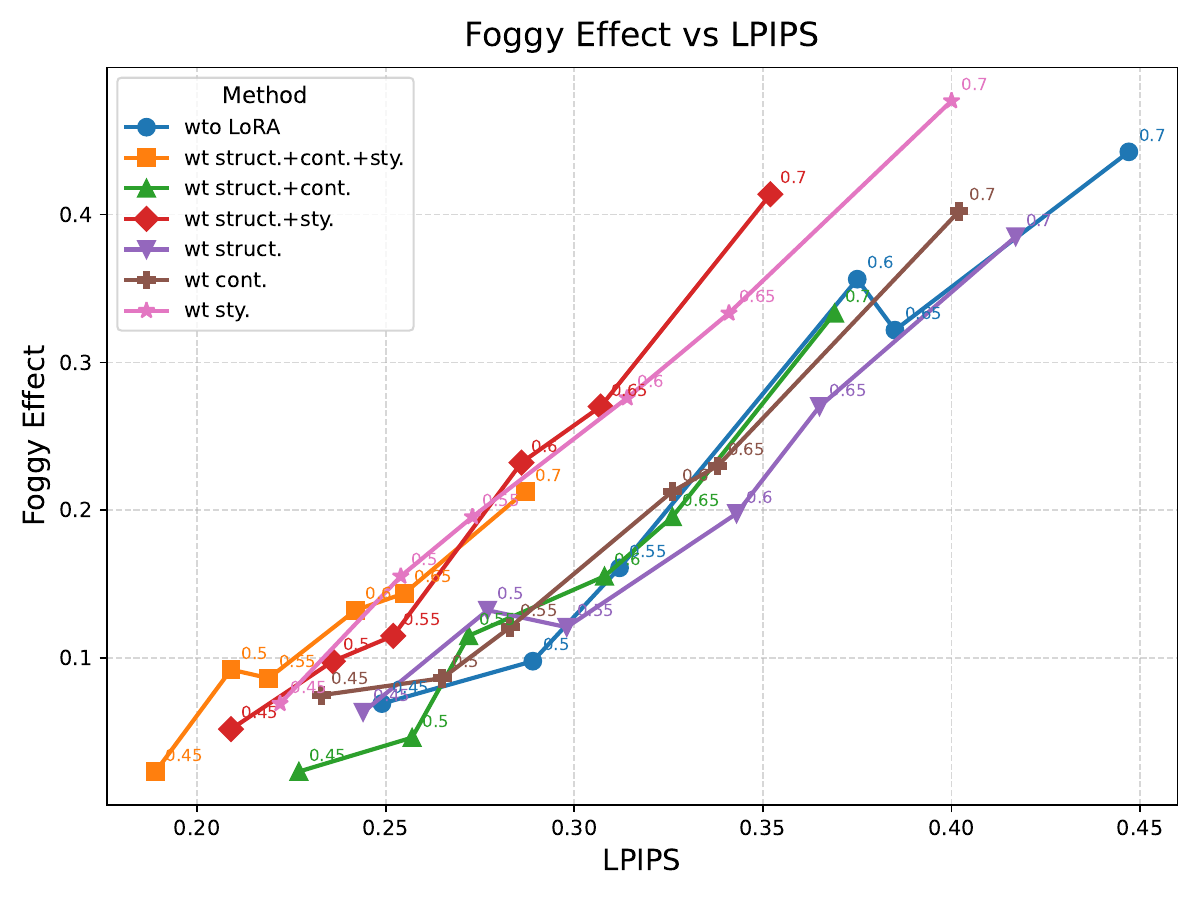}}
    
    \caption{Quantitative results regarding  structure preservation and image realism  vs. weather effect.}
    \label{fig:ssim_lpips111}
\end{figure}

We present the results of image generation quality in~\cref{fig:ssim_lpips111}, as we evaluated structure preservation, image realism, and the effectiveness of the weather effect. Annotations in the figure refer to strength values. Generation without LoRA yields the best overall weather-synthesis effect, except for \texttt{cloudy} and \texttt{foggy} conditions. However, it shows lower SSIM than using a LoRA-enabled model at the same strength setting, indicating worse structure preservation. Using only text prompts without LoRA weights trained for structure preservation indeed limits the controllability of adding weather effects to a given scene and its applicability for systematic robustness tests of downstream tasks. The LPIPS values obtained by SDXL without LoRA are generally higher than those with LoRA at the same strength setting, indicating that the generated images are statistically less perceptually similar, i.e., less realistic. Among the results with different LoRA weights, images generated with our structure-LoRA weights show a greater degree of weather-effect manipulation. Compared with images generated using content-LoRA weights, those generated with structure-LoRA weights produce better weather synthesis, with a slight loss in structure preservation. This might be that content-LoRA, which represents object information, also contains partial structural information, but its use might limit style/weather manipulation. The weather synthesis success rates when using style-LoRA for cloudy and foggy conditions are generally higher than those using only structure-LoRA. We inspect the original images and find that many of them exhibit a mild, cloudy, or foggy effect, which has been encoded in style-LoRA. But the realism of the generated images using style-LoRA is much lower (see the high LPIPS value when strength=0.7 for cloudy effect synthesis). Combining style and structure LoRAs results in better cloudy and foggy synthesis effect while retaining structure information.

\begin{figure}[!t]
    \centering
    \includegraphics[width=.9\linewidth]{ 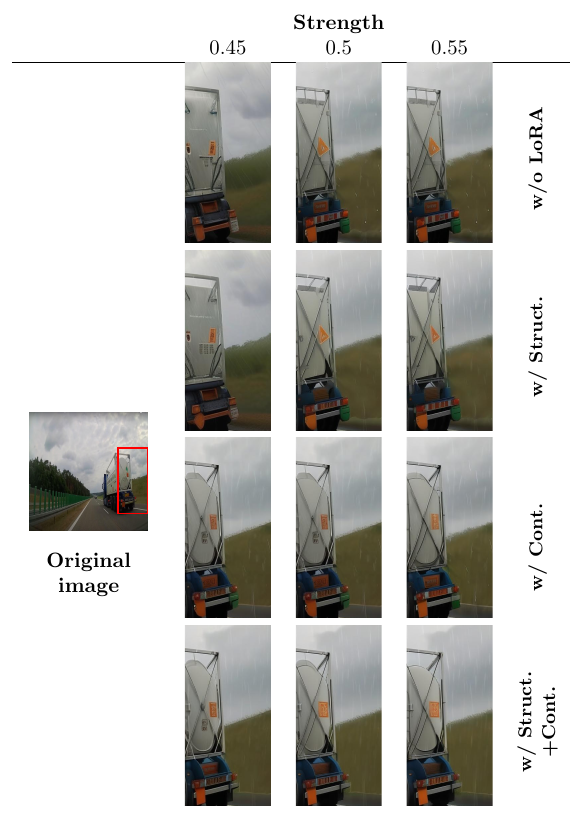}
    \caption{I2I results using  strength $= 0.45, 0.5, 0.55 $, with  LoRA setting: w/o  LoRA , w/ stlye, w/ content, w/ style+content LoRA, respectively.  Structure-LoRA preserves structure while enable more obvious weather synthesis effect guided by text prompt.   }
    \label{fig:strength_qual}
\end{figure}

\begin{figure}[t]
    \centering
    \includegraphics[width=.9\linewidth]{ 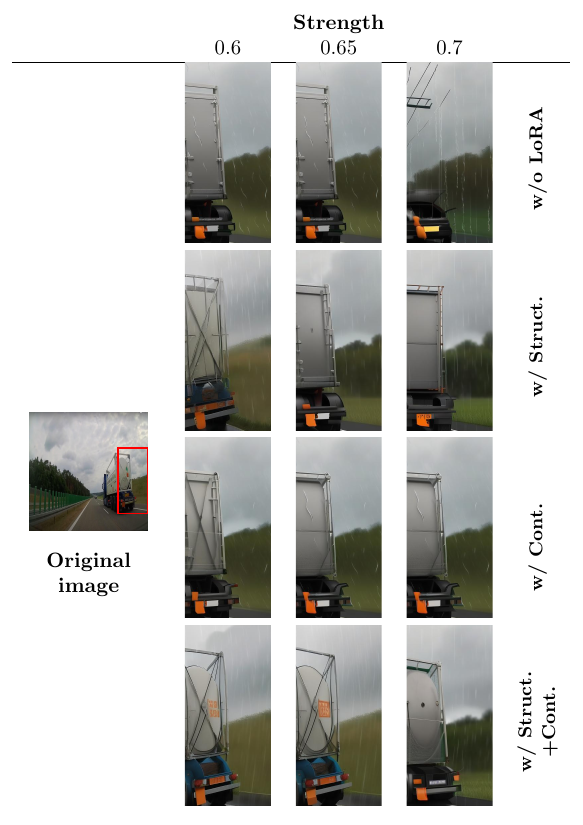}
    \caption{I2I results using  strength $= 0.6, 0.65, 0.7 $, with  LoRA setting: w/o  LoRA , w/ stlye, w/ content, w/ style+content LoRA, respectively.  Structure-LoRA preserves structure while enable more obvious weather synthesis effect guided by text prompt.   }
    \label{fig:strength_qual222}
\end{figure}

\subsection{Qualitative results}

\begin{figure}[!t]
    \centering
    \includegraphics[width=.85\linewidth]{ 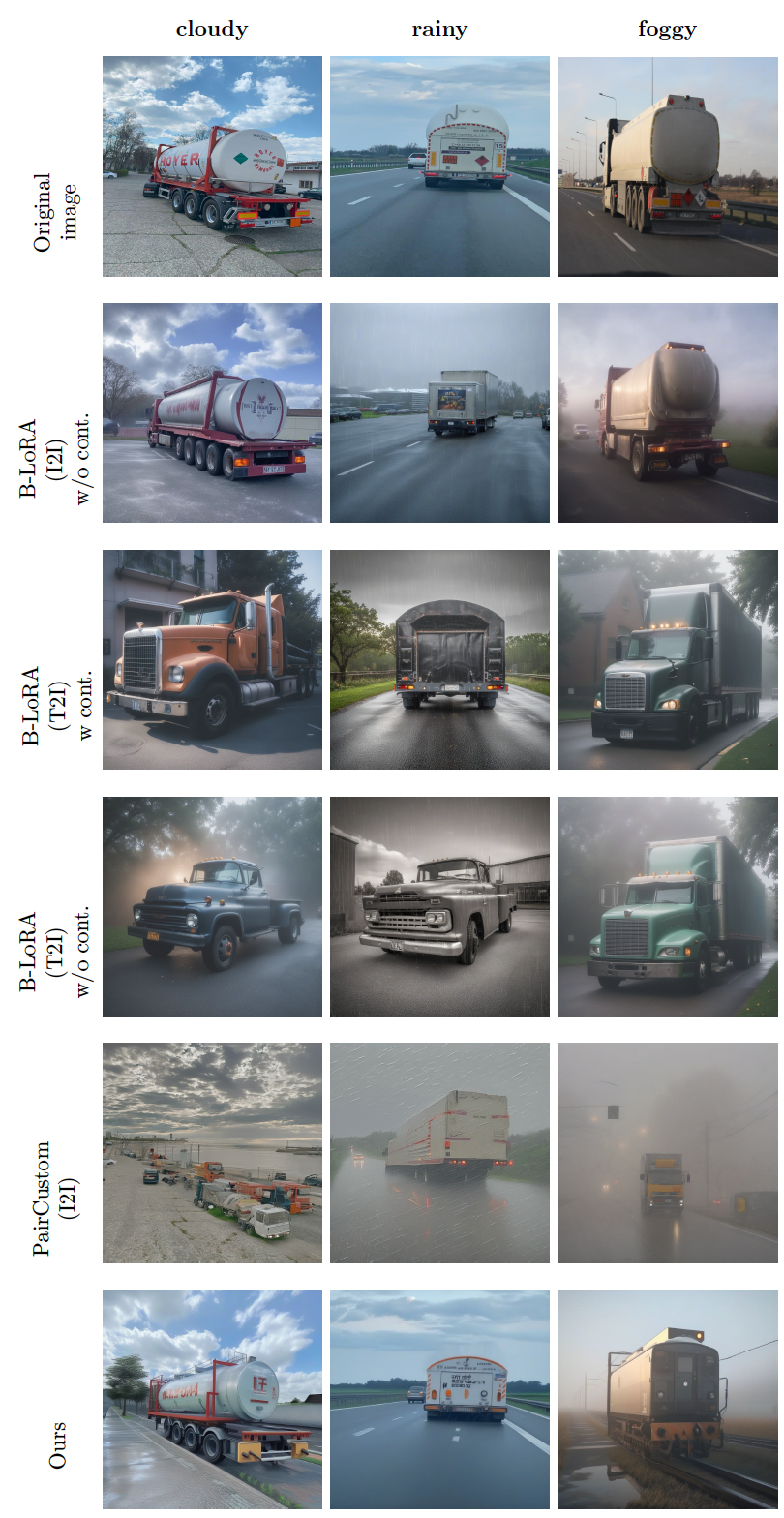}
    \caption{Comparison with B-LoRA and PairCustomization. For B-LoRA, we learn  style-LoRA  from a synthetic pair of  sunny-[cloudy/rainy/foggy] images.  We generate images using both I2I and T2I. As text provides limited structural information, we provide content-LoRA (which contains partial structural information)  for T2I setup for a fairer comparison. For PairCustomization, we use I2I generation with the learned style-LoRA  from the reference image pair.  }
    \label{fig:compar_qua}
\end{figure}

In~\cref{fig:strength_qual,fig:strength_qual222}, we show rain synthesized images using different combinations of LoRA  (w/ structure, w/ content, and w/ structure plus content) or without using LoRA. Images generated with lower strength values have almost no observable synthesized  weather effect, while a higher value results in synthesized rain streak and reflection on the road. The usage of structure-LoRA  preserves most structure information, compared to that without using LoRA  (generating images under strength=0.7 configuration replaces the original truck by a car).  Observed from the results of LoRA w/ content  and LoRA w/ structure plus content, the structure-LoRA  better preserves overall structure details of the original image under high strength configuration while allowing for more obvious rainy synthesis effect. Images of other synthesized weather effects are in~\cref{fig:compar_qua}.

\noindent \textbf{Comparisons with baselines.} 
\begin{table*}[!t]
    \centering
    \scriptsize
    \resizebox{0.8\linewidth}{!}{%
    \begin{tabular}{lccccccc}
        \toprule
       
        \textbf{Method}  & {\textbf{LPIPS}$\downarrow$} 
        & {\textbf{SSIM}$\uparrow$} 
        & {\textbf{Rainy}$\uparrow$} & {\textbf{Cloudy}$\uparrow$} & {\textbf{Snowy}$\uparrow$} & {\textbf{Foggy}$\uparrow$} & Avg. 
      \\
        \midrule
        Pair (I2I)                     & 0.541 & 0.547 & \textbf{95.38} &  12.14  &  \textbf{95.95}  & 29.48 & \textbf{58.24}  \\
        BLoRA (T2I) w/ C                & 0.726 & 0.318 & 60.92 & \underline{77.59}  &   0  &  20.69  & 39.8 \\
        BLoRA (T2I) w/o C               & 0.738 & 0.316 & 47.7 &  \textbf{100}  & 0.57   & 0  & 37.07 \\
        BLoRA (I2I) w/o C               & \textbf{0.431} & \underline{0.563} & 53.45 &   39.66 & 2.3   & \textbf{43.1} & 34.63  \\
        Structure LoRA                  & \underline{0.388} & \textbf{0.601} & \underline{76.44} &  28.74  & \underline{72.83}   &  \underline{38.51} & \underline{54.13} \\
        \bottomrule
    \end{tabular}%
    }
    \caption{Performance of different methods: we report LPIPS, SSIM, and weather synthesis success rate. Lower LPIPS and higher SSIM indicate better quality, while higher weather synthesis success rate indicates better weather manipulation. 
    }
    \label{tab:comparison results}
\end{table*}

The quantitative comparisons between our method with B-LoRA and PairCustomization are presented in~\cref{tab:comparison results}. Our method balances the synthesis of weather effects with the preservation of the original structure of the input images. We achieve the second-best weather-synthesis performance, except for cloudy weather, while our method preserves the most structural information of the original images (highest SSIM). The realism of our generated images ranks second compared with that of B-LoRA (I2I) w/o C.

We qualitatively compare our weather synthesized images with those generated using B-LoRA and PairCustomization in~\cref{fig:compar_qua}. Under I2I generation setting, our method maintains more structure information, while adding more realistic weather effects to the images. B-LoRA instead results in almost no synthesized weather effect, and PairCustomization completely changes the structure information of the original images. Using B-LoRA T2I generation option, the generated images demonstrate different structure and content information even using the content LoRA   of the original image. These results show that our method preserves more structure information with less compensation regarding the magnitude of style manipulation.

\subsection{Limitations}
Although structure-LoRA preserves more structural information of images during I2I translation, not all the details of the original images are maintained. For example, the number plate and hazmat sign in the images are important for identifying a truck and the hazardous material being transported. To recover these details, incorporating text inpainting could be a reasonable approach. Moreover, we observe that if an input image contains contrasting weather condition to the target weather condition, the model might fail to synthesize the targeted weather effect. This might be attributed to the insufficient guidance of a given text prompt to cover the original weather effect when the semantic gap between them is large. For future work, disentangling explicitly the weather effect from the original image content would be a promising direction. 

\section{Conclusions}
Through an architectural analysis of SDXL-ControlNet, we uncover that structural condition extracted by the middle block of ControlNet is highly correlated to structure preservation in image generation. We propose a weather synthesis method that learns the corresponding LoRA matrices to image structure, enabling the manipulation of weather effects while retaining the original image structure and content. Our method requires light computations and training time, and improves weather effect variety.

\bibliographystyle{ieeenat_fullname}
\bibliography{main}


\end{document}